\documentclass[10pt,twocolumn,letterpaper]{article}

\usepackage[pagenumbers]{cvpr} 

\usepackage{graphicx}
\usepackage{amsmath}
\usepackage{amssymb}
\usepackage{booktabs}
\usepackage{paralist}

\usepackage[pagebackref,breaklinks,colorlinks]{hyperref}

\usepackage[capitalize]{cleveref}
\crefname{section}{Sec.}{Secs.}
\Crefname{section}{Section}{Sections}
\Crefname{table}{Table}{Tables}
\crefname{table}{Tab.}{Tabs.}

\def\cvprPaperID{*****} 
\def\confName{AI535}
\def\confYear{S2022}

\begin{document}

\title{Contrastive Learning for OOD in Object detection}

\author{Rishab Balasubramanian \hspace{20pt} Rupashree Dey \hspace{20 pt} Kunal Rathore\\
Oregon State University\\
{\tt\small \{balasuri, deyr, rathorek\}@oregonstate.edu}
}
\maketitle





\begin{abstract}
    Contrastive learning is commonly applied to self-supervised learning, and has been shown to outperform traditional approaches such as the triplet loss and N-pair loss. However, the requirement of large batch sizes and memory banks has made it difficult and slow to train. Recently, Supervised Contrasative approaches have been developed to overcome these problems. They focus more on learning a good representation for each class individually, or between a cluster of classes. In this work we attempt to rank classes based on similarity using a user-defined ranking, to learn an efficient representation between all classes. We observe how incorporating human bias into the learning process could improve learning representations in the parameter space. We show that our results are comparable to Supervised Contrastive Learning for image classification and object detection, and discuss it's shortcomings in OOD Detection. All code available at \texttt{\url{https://github.com/rishabbala/Contrastive_Learning_For_Object_Detection}}
\end{abstract}


\section{Introduction}
\label{sec:intro}

Learning ``good" representations for downstream tasks has been a long sought after task in computer vision. The most popular approach for representation learning are dicriminative approaches, where a model is trained with a task-defined loss to differentiate between the known classes. This attempts to create a linear separation between the class boundaries in the higher dimensional parameter space. However no explicit emphasis is placed on learning (dis)similarity between classes, which causes the learned projections to be in such a manner that they can be easily clustered, and linearly separable with no focus on the underlying embeddings. Creating representations of objects in such a way that they incorporate a notion of ``closeness" with other similar, but not same objects is principal in representation learning. For example, we would want the images of ``cars" to be ``closer" in some sense to ``trucks" than to ``dogs".\\

One recent trend for incorporating the notion of ``closeness" into learning object representations is using contrastive learning. Contrastive Learning follows from traditional triplet-loss, where similarity between an ``anchor" and ``positive" is maximized, and the similarity between ``anchor" and ``negative" is minimized. Starting off as a method of self-supervised learning (\cite{chen2020simple}, \cite{he2020momentum}, \cite{chen2020improved}), contrastive learning has also been extended to supervised learning \cite{khosla2020supervised}. In this work we attempt to incorporate human bias and domain knowledge into the training regime through the form of ranking classes based on similarity with other classes. We evaluate our proposed method on image classification and object detection and show that we achieve comparable scores to that of \cite{khosla2020supervised}, and much better than discriminative methods. We also test our model on OOD detection and discuss the shortcomings.\\

\section{Related Work}

A popular choice of loss function for training neural networks are the L1/L2 losses for regression, and cross-entropy losses for classifications. Although these methods have achieved state of the art results on various tasks, they do not emphasize learning a good representation of the data. The training regime pushes the model to learn features for different classes that are eventually linearly separable.\\

\begin{figure*}[!ht]
    \centering
    \subfloat[\centering 1 positive class]{{\includegraphics[width=0.3\textwidth, height=5cm]{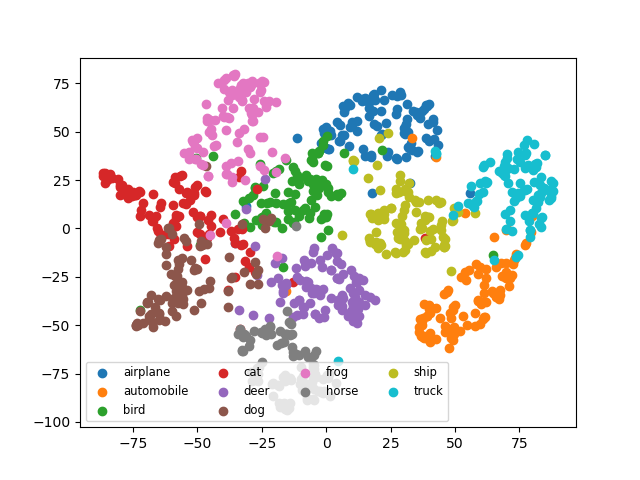} }}
    \subfloat[\centering 5 positive class]{{\includegraphics[width=0.3\textwidth, height=5cm]{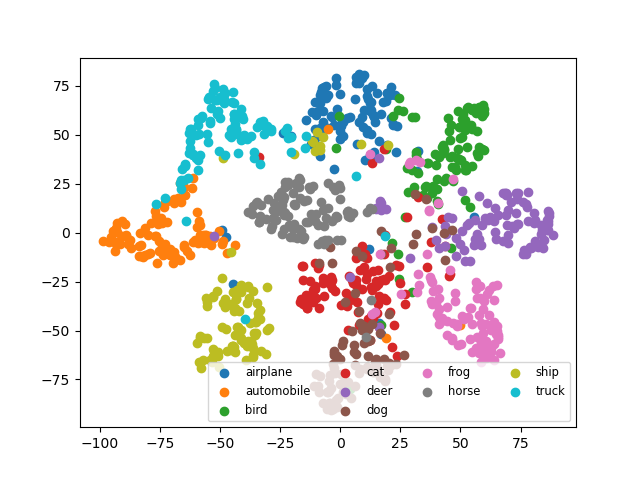} }}
    \subfloat[\centering baseline]{{\includegraphics[width=0.3\textwidth, height=5cm]{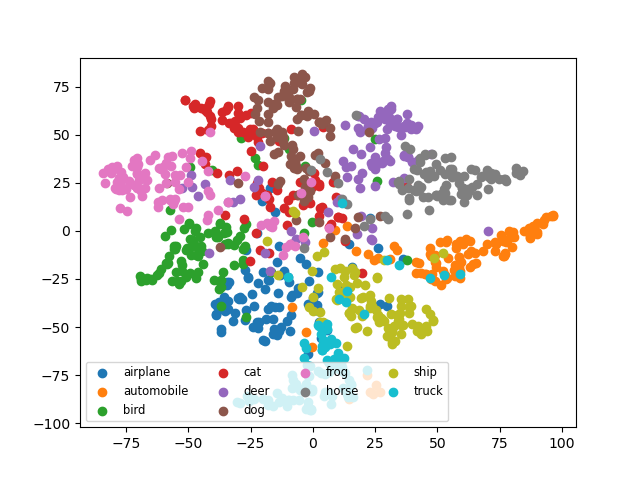} }}\\
\caption{T-SNE plots for CIFAR10}
\label{fig:tsne}
\end{figure*}

As discriminative approaches place no emphasis on learning a ``distance" metric, and generative approaches are too computationally expensive for simple tasks, there has been a shift towards using contrasive learning as a method for enforcing better representation learning. Contrastive learning uses a similarity inspired loss to incorporate ``distance" into the training process. Self-supervised contrastive learning approaches create two transformations of the image representing the ``anchor" and ``positive", and consider all other images as ``negative". However, as expected they require large batch sizes and memory to learn useful representations \cite{chen2020simple}, \cite{he2020momentum}, \cite{chen2020improved}. Supervised Contrastive learning \cite{khosla2020supervised} overcomes these challenges by using all other images in the same class as ``positives", and images in other classes as negative. This does not however take into account the relationship between different classes, which could be used to learn a better representation. RINCE \cite{hoffmann2022ranking} builds upon this by clustering similar classes together, and using the information to learn better representations. In this work, we extend \cite{hoffmann2022ranking} to incorporate custom ranking, where the classes are ranked based on a user defined scheme, and we observe the impact of incorporating user bias into the training process. \\

To verify the efficiency of our learned representations, we test our approach not only on image classification and object detection, but OOD detection as well. Some of the most common methods for OOD detection use discriminative training. One popular approach is using a threshold on the class prediction probabilities, and declare an input as OOD if all predictions are below the given threshold. Another well-known approach tackles OOD detection at the input side, by passing the inputs through a one-vs-many SVM to determine whether the input is inside/outside the distribution. One could also gather exrta data of random objects to train a new classifier to detect OOD images at the input stage. In this work we will compare our results against the first method of using a threshold.\\

\section{Methodology}

We create a custom ranking system that incorporates human bias into the model. Specifically, for each class we list four other classes that are similar to it. The anchor is then compared with its second augmentation, other images in the same class, and the ranked classes using a one-vs-all contrastive loss.\\

Formally, for each anchor (query) image $q$, we rank a number of similar classes as $\mathcal{P}_1 \cdots \mathcal{P}_r$, where $r$ denotes the number of positive classes in our ranking. We also define a negative class $\mathcal{N}$, as the set of all other image classes. Let $h(q,x)$ be the cosine similarity between the query and any other image $x$. Then, we can define our objective as being able to train our model such that:

\begin{equation}
h(q, \mathcal{P}_1) > h(q, \mathcal{P}_2) > \cdots h(q, \mathcal{P}_r) > h(q, \mathcal{N})
\end{equation}

We do this by defining a loss $L = \sum_{i=1}^{r} l_{i}$ where

\begin{equation}
    \mathit{l_{i}} = -\log \frac{\sum_{p \in \mathcal{P}_{i}} \exp(\frac{h(q,p)}{\tau_{i}})}{\sum_{p \in \cup_{j \geq i} \mathcal{P}_{i}} \exp(\frac{h(q,p)}{\tau_{i}}) + \sum_{n \in \mathcal{N}} \exp(\frac{h(q,n)}{\tau_{i}})}    
\end{equation}

This can be thought of as recursively computing the loss ($L$), when considering the current highest ranked class ($i$) as ``positive" and all other classes as negative. After computing the loss, the current highest ranked class ($i$) is removed, and the loss is computed again for class $i+1$. To ensure good separation, we set $\tau_{i+1} > \tau_{i}$, following the empirical studies provided in \cite{hoffmann2022ranking}.\\

Opposed to \cite{hoffmann2022ranking}, we rank classes instead of clustering them into groups. The difference between them is that in our ranking, any class could be ranked similar to any other class, with a user defined score. However, in clustering as done in \cite{hoffmann2022ranking}, only the classes within the same cluster are considered similar. For example, \cite{hoffmann2022ranking} puts the classes ``aeroplane" and ``ship" together as ``vehicles". However, from human knowledge, we know that an ``aeroplane" is also (probably more) similar to a ``bird" than a ``ship". Hence in our method, we create the ranking for the class ``aeroplane" as \{``bird", ``ship", ... \} with decreasing order of similarity from left to right.\\

We train and observe the model's performance on image classification, object detection, and out-of-distribution (OOD) detection.\\

\section{Results}
We compare our results with SupCL (\cite{khosla2020supervised} where there is no ranking), RINCE (\cite{hoffmann2022ranking}, where similar classes are clustered together), and SoftMax (common discriminative approach of training a ResNet50 with a SoftMax loss). To maintain fairness in evaluations, we use a ResNet50 backbone for all our models, train image classification models for 300 epochs, and object detection models for 500 epochs with the same training methods.\\

\subsection{Image Classification}
We first evaluate our method on image classification. We train and test our model on the CIFAR10 dataset. Fig \ref{fig:tsne} shows the TSNE plots of projecting the learned representations when trained on CIFAR10. We observe that the discriminative learning approach learns a cluttered object representation, where the projected distance does not represent a good ``closeness" in terms of class similarity. When ($r=1$), we see that the projected representations are clustered into well defined classes. When we increase the ranking to $r=3$ we see that the learned representations show expected ``closeness" in the projected space, where the animals are further away from the vehicles, but similar classes  like ``bird", and ``cat" are still close together. Table \ref{table:CIFAR10} shows the classification accuracy on CIFAR10 dataset, after training all three models in a similar fashion for 300 epochs. Again we see that our model performs comparable to supervised contrastive learning and is much better than discriminative learning.\\

\begin{table}[h]
  \centering
  \begin{tabular}{@{}lc@{}}
    \toprule
    Method & Classification Accuracy \\
    \midrule
    SupCL ($r$=1) & 0.9085\\
    Ours ($r$=3) & 0.8937\\
    SoftMax & 0.811\\
     \hline
    \end{tabular}
     \caption{Classification Accuracy on CIFAR10}
     \label{table:CIFAR10}
\end{table}

\begin{table}[h]
  \centering
  \begin{tabular}{@{}lc@{}}
    \toprule
    Method & Classification Accuracy \\
    \midrule
    SupCL ($r$=1) & 0.6499\\
    Ours($r$=3) & 0.6068\\
    RINCE($r$=5) & 0.6368\\
    SoftMax & 0.5829\\
    \bottomrule
  \end{tabular}
   \caption{Classification Accuracy on VOC2007}
 \label{table:VOC}
\end{table}

\begin{table*}[t]
\resizebox{\textwidth}{!}{%
\begin{tabular}{ p{8cm} p{2cm} p{2cm} p{2cm}  }
 \hline
 Model & AP & AP50 & AP75 \\
 \hline
Faster RCNN (trained on VOC2007 train+val) ResNet50 FPN & 45.254 & 72.746 & 49.338\\
\hline
\end{tabular}}
\caption{Object Detection scores}
\label{table:map}
\end{table*}

\subsection{Object Detection}

We first evaluate the efficiency of the Faster-RCNN model in predicting bounding boxes. To do so, we compute the AP scores using a pre-trained model. Table \ref{table:map} shows the AP, AP50, and AP75 scores of the Object detection module we used. This is lower than the values reported in the Faster RCNN paper, and also lower than more modern approaches. Since the objective was not only
object detection, we did not try different detection models.\\

\begin{figure*}[h!]
    \centering
    \subfloat{{\includegraphics[width=0.3\textwidth, height=5cm]{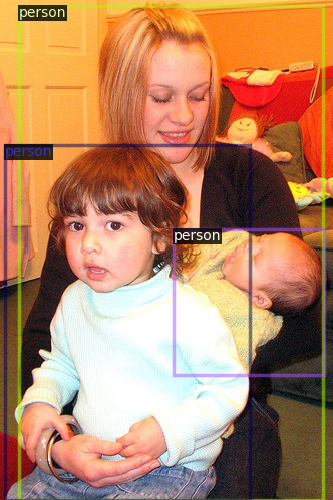} }}
    \subfloat{{\includegraphics[width=0.3\textwidth, height=5cm]{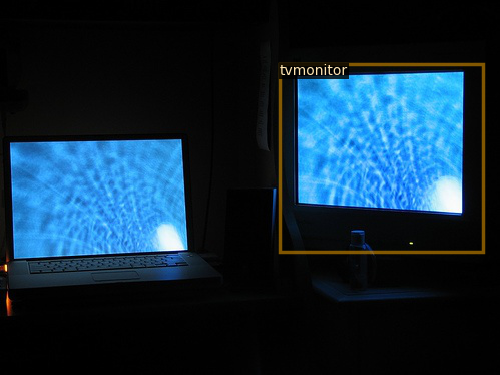} }}
    \subfloat{{\includegraphics[width=0.3\textwidth, height=5cm]{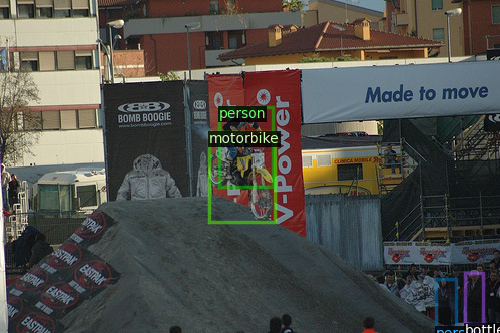} }}\\
    \subfloat{{\includegraphics[width=0.3\textwidth, height=5cm]{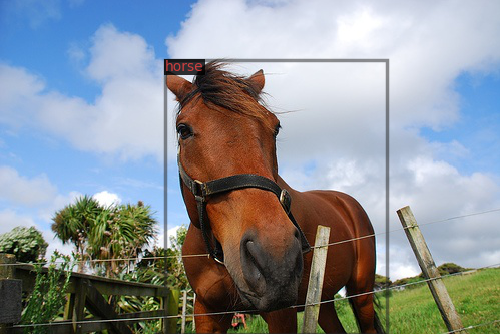} }}
    \subfloat{{\includegraphics[width=0.3\textwidth, height=5cm]{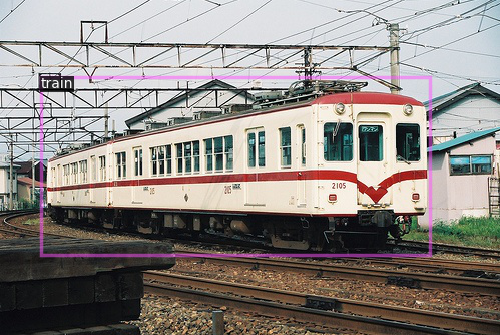} }}
    \subfloat{{\includegraphics[width=0.3\textwidth, height=5cm]{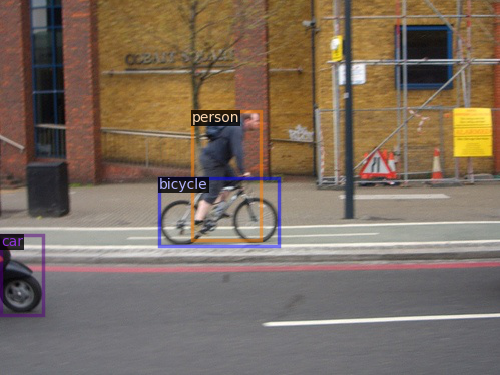} }}\\
\caption{ Object detection Results}
\label{fig:results}
\end{figure*}

Table \ref{table:VOC} shows the classification accuracy on the VOC2007 dataset, and Fig \ref{fig:results} shows the results from our model. We observe that again our accuracy is comparable to RINCE \cite{hoffmann2022ranking} and SupCL \cite{khosla2020supervised}, while the discriminative classification recieves a much lower score. We however observe that our scores are slightly lesser than SupCL and RINCE. As all other parameters were similar during testing, the factor that affects these results the most is ranking, and the user-tuned class-similarity scores. As most classes are completely different to each other in the dataset, it resulted in sparser ranking which affects the performance here.\\


\begin{figure*}[h!]
    \centering
    \subfloat[\centering baseline]{{\includegraphics[width=0.3\textwidth, height=5cm]{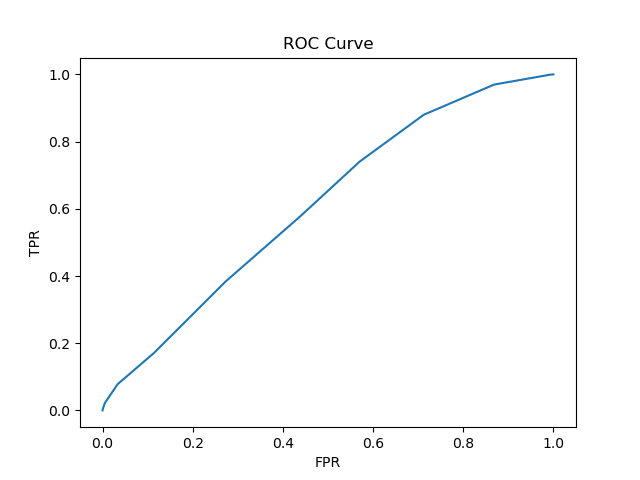} }}
    \subfloat[\centering 1 positive class ($r=1)$]{{\includegraphics[width=0.3\textwidth, height=5cm]{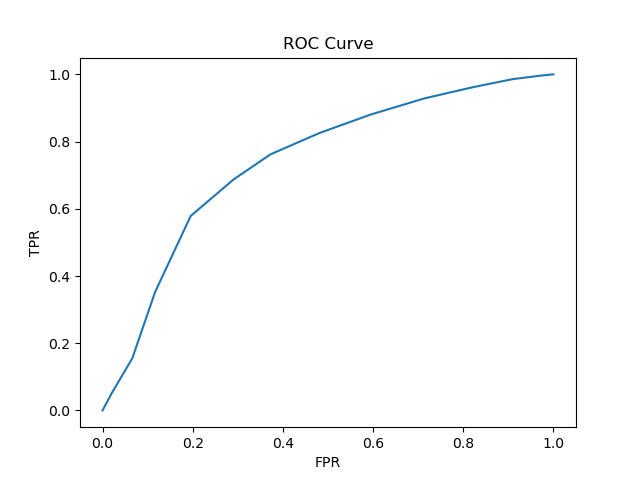} }}
    \subfloat[\centering 5 positive classes ($r=5)$]{{\includegraphics[width=0.3\textwidth, height=5cm]{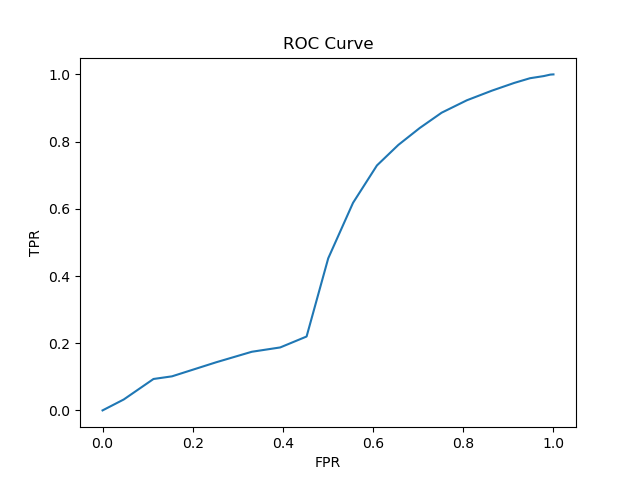} }}\\
\caption{ROC plots for CIFAR10}
\label{fig:rocCIFAR10}
\end{figure*}

\begin{figure*}[h!]
    \centering
    \subfloat[\centering baseline]{{\includegraphics[width=0.3\textwidth, height=5cm]{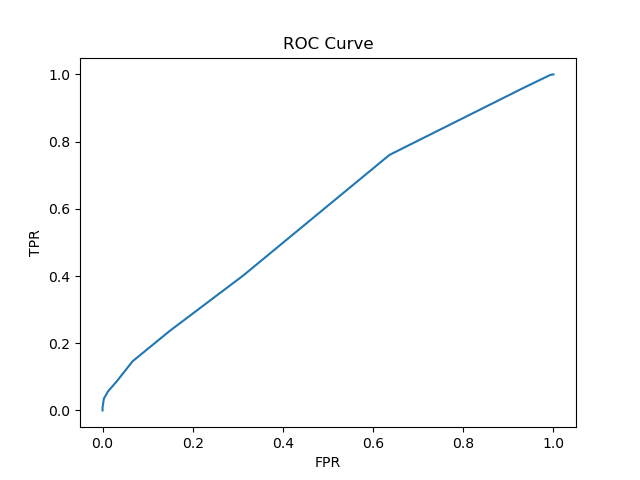} }}
    \subfloat[\centering 1 positive class ($r=1)$]{{\includegraphics[width=0.3\textwidth, height=5cm]{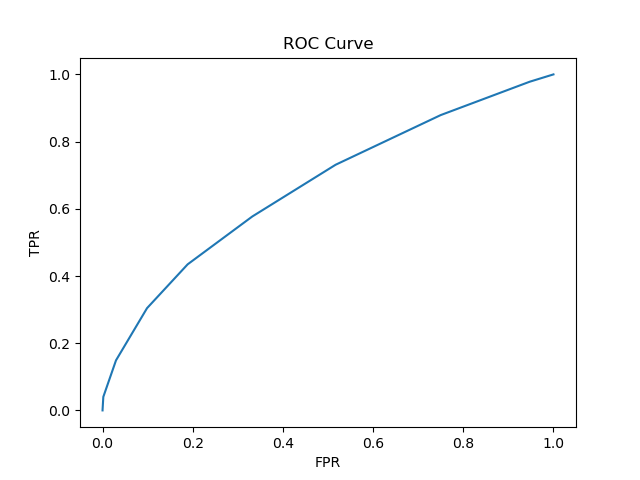} }}
    \subfloat[\centering 5 positive classes ($r=5)$]{{\includegraphics[width=0.3\textwidth, height=5cm]{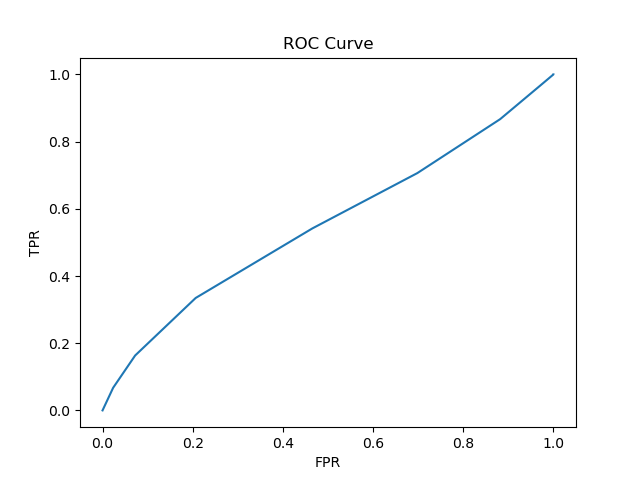} }}\\
\caption{ROC plots for VOC}
\label{fig:rocVOC}
\end{figure*}

\subsection{OOD Detection}

We finally evaluate our model's performance for OOD image classification, and OOD image detection. First we test our model on CIFAR10 with 2 classes withheld. The ROC curves are shown in Fig \ref{fig:rocCIFAR10}, and the AUROC in Table \ref{table:CIFAR10OOD}. We see that we perform very poorly in comparison to the other two baselines. We also evaluate on the VOC2007 dataset, with 2 classes withheld again, and show the ROC curve in Fig \ref{fig:rocVOC} and the AUROC in Table \ref{table:VOCOOD}. We again see that our model do not perform better than the baselines. This shows that our method enforces good representation learning when human input is given, but the representations for new classes are poor. This can probably be attributed to the fact that any new input is pushed too close to one of the known classes.\\

It would be interesting to try a slight variation of these experiments with a reference feature vector. We first collect some additional images of classes that are not in the known dataset. Then we define a reference feature vector, for example $f = \{1, 1, \cdots 1\}$. Now we train the model again, and for each object which is known to be OOD, we rank $f$ as its only similar class. To make a fair evaluation, we could withhold 4 classes, of which only 2 are used in training. However, due to time limitations, we were not able to run this experiment.\\

\begin{table}[h!]
  \centering
  \begin{tabular}{@{}lc@{}}
    \toprule
    Method & AUROC \\
    \midrule
    SupCL ($r$=1) & 0.7426\\
    Ours($r$=3) & 0.4911\\
    SoftMax & 0.611\\
     \hline
    \end{tabular}
     \caption{AUROC on CIFAR10 with 2 classes witheld}
     \label{table:CIFAR10OOD}
\end{table}

\begin{table}[h!]
  \centering
  \begin{tabular}{@{}lc@{}}
    \toprule
    Method & AUROC \\
    \midrule
    SupCL ($r$=1) & 0.6679\\
    Ours($r$=5) & 0.5532\\
    SoftMax & 0.5621\\
 \hline
\end{tabular}
 \caption{AUROC on VOC2007 with 2 classes witheld}
 \label{table:VOCOOD}
\end{table}

\section{Conclusion}
Contrastive Learning has become an improved and established technique for better feature learning in deep neural networks. In this work, we have learned how Contrastive Learning works, its types, components and implementations, we have done experimentation with a belief of better representation learning for object detection and OOD handling. We trained and evaluated our method of human-in-the-loop contrastive learning and current state of the art methods and compared the results. As per the experiments, our approach performed comparable to the existing methods in classification tasks when all classes were known, but fails in OOD detection. We also provide a method for future improvements to our work.

{\small
\bibliographystyle{ieee_fullname}
\bibliography{egbib}
}

\end{document}